\documentclass{article}

\usepackage{microtype}
\usepackage{graphicx}
\usepackage{subfigure}
\usepackage{booktabs} 

\usepackage{hyperref}

\usepackage[accepted]{icml2021}

\usepackage{amsmath,amsfonts,bm}

\def\eqref#1{equation~\ref{#1}}

\def\1{\bm{1}}

\def\vx{{\bm{x}}}

\DeclareMathAlphabet{\mathsfit}{\encodingdefault}{\sfdefault}{m}{sl}
\SetMathAlphabet{\mathsfit}{bold}{\encodingdefault}{\sfdefault}{bx}{n}

\newcommand{\R}{\mathbb{R}}

\usepackage{url}
\usepackage{multirow}
\usepackage{bbding}

\definecolor{dkred}{rgb}{0.5,0,0}
\definecolor{dkgreen}{rgb}{0,0.6,0}
\definecolor{gray}{rgb}{0.5,0.5,0.5}
\definecolor{mauve}{rgb}{0.58,0,0.82}

\newcommand{\mr}[2]{\multirow{#1}{*}{#2}}
\newcommand{\mc}[3]{\multicolumn{#1}{#2}{#3}}
\definecolor{customgray}{rgb}{0.3,0.3,0.3}
\definecolor{customgreen}{RGB}{140,211,89}
\newcommand{\std}[1]{\textcolor{customgray}{\scriptsize{$\pm$#1}}}
\newcommand{\duihao}{\textcolor{dkgreen}{\CheckmarkBold}}

\icmltitlerunning{R-GSN: The Relation-based Graph Similar Network for Heterogeneous Graph}

\begin{document}

\twocolumn[
\icmltitle{R-GSN: The Relation-based \\ Graph Similar Network for Heterogeneous Graph}



\icmlsetsymbol{equal}{*}

\begin{icmlauthorlist}
\icmlauthor{Xinliang Wu}{xjtu}
\icmlauthor{Mengying Jiang}{xjtu}
\icmlauthor{Guizhong Liu}{xjtu}
\end{icmlauthorlist}

\icmlaffiliation{xjtu}{School of Information and Communications, Xi'an Jiaotong University, Xi’an, China}

\icmlcorrespondingauthor{Guizhong Liu}{liugz@xjtu.edu.cn}

\icmlkeywords{Graph neural networks, Heterogeneous Graph, ogbn-mag}

\vskip 0.3in
]



\printAffiliationsAndNotice{}

\begin{abstract}
Heterogeneous graph is a kind of data structure widely existing in real life. Nowadays, 
the research of graph neural network on heterogeneous graph has become more and more popular. 
The existing heterogeneous graph neural network algorithms mainly have two ideas, 
one is based on meta-path and the other is not. The idea based on meta-path often requires 
a lot of manual preprocessing, at the same time it is difficult to extend to large scale graphs. 
In this paper, we proposed the general heterogeneous message passing paradigm and designed R-GSN that does not need meta-path, which is much improved compared to the baseline R-GCN. Experiments have shown that our R-GSN algorithm 
achieves the state-of-the-art performance on the $\texttt{ogbn-mag}$ large scale heterogeneous graph dataset.
\end{abstract}
\section{Introduction}
In practical applications, many data can be regarded as graphs, such as social networks, academic networks and knowledge graphs. There are three types of common tasks on the graph, namely the node classification, the link prediction and the graph classification. In recent years, due to the numerous application scenarios and challenges of graphs, the research of graph neural networks has become more and more popular, and many classic algorithms on the homogeneous graphs have been proposed, such as ChebNet~\citep{defferrard2016convolutional}, GCN~\citep{kipf2016semi}, GraphSAGE~\citep{hamilton2017inductive}, GAT~\citep{velivckovic2017graph}, GIN~\citep{xu2018powerful}, DeepGCNs~\citep{li2019deepgcns} and GeomGCN~\citep{pei2020geom}. 

But obviously in real scenes many graphs are heterogeneous. Therefore, researchers began to pay more attention to  neural networks on heterogeneous graphs. R-GCN~\citep{schlichtkrull2018modeling} is a heterogeneous graph neural network designed for knowledge graphs, which solved the tasks of entity classification and link prediction on the knowledge graphs; note that R-GCN aggregates the first-order neighborhoods. In heterogeneous graphs, by predefining meta-path~\citep{dong2017metapath2vec}, the semantic high-order neighbors of the target node can be found and utilized. HAN~\citep{wang2019heterogeneous} used meta-path to aggregate high-order neighbors and introduced attention mechanism. MAGNN~\citep{fu2020magnn} is similar to HAN, the difference is that the features of all the nodes on the meta-path have been used in MAGNN. GTN~\citep{yun2019graph} is a novel model that can automatically learn the meta-path without manually predefining it in advance. HGT~\citep{hu2020heterogeneous} got rid of the meta-path, adopted meta-relation, and designed an aggregation method similar to Transformer~\citep{vaswani2017attention}.

Since most of the existing algorithms are based on the meta-path, they require a lot of manual processing and have poor generalization. For this we propose an end-to-end R-GSN algorithm based on R-GCN, which does not need meta-path and can be easily applied to large scale heterogeneous graph datasets, such as $\texttt{ogbn-mag}$~\citep{hu2020open}. Our contributions are as follows:
\begin{itemize}
	\item We propose a general heterogeneous message passing paradigm. Under this framework, heterogeneous message passing is summarized into the following four steps: Message Transform, Intra-relation Aggregation,  Inter-relation Aggregation and Status Update. R-GCN~\citep{schlichtkrull2018modeling} is a special instance of this paradigm.
	\item We propose the R-GSN algorithm, which conforms to the general heterogeneous message passing paradigm, and we design two novel similarity aggregation methods, called the SIM-ATTN aggregation and the SIM aggregation; note that the SIM-ATTN aggregation includes an attention mechanism while the SIM aggregation does not.
	\item In R-GSN, we use normalization operations in several places. Experiments prove the necessity of normalization operations. there are two other operations that are also useful, one is the adversarial training method called FLAG~\citep{kong2020flag}, and the other is specifically for the $\texttt{ogbn-mag}$ dataset in generating parameterized initial features for nodes without initial features through pre-propagation.
	\item R-GSN achieves the state-of-the-art performance on the $\texttt{ogbn-mag}$ dataset. The average accuracy on the validation dataset and test dataset are respectivly 51.82\% and 50.32\%, both are higher than those of the existing algorithms. At the same time, compared with the baseline R-GCN, there is a 4.21\% gain on the validation dataset and a 3.54\% gain on the test dataset.
	\item We extend R-GSN and propose a homogeneous version algorithm called GSN and a meta-path version algorithm called M-GSN. Respectively, we conducte experiments on $\texttt{ogbn-arxiv}$, $\texttt{IMDB}$ and $\texttt{DBLP}$ datasets, and prove the effectiveness of the aggregation method proposed in this paper.
\end{itemize}


\section{Perliminaries}
\subsection{Heterogeneous Graph}
A heterogeneous graph is defined as a graph $\mathcal{G}=(\mathcal{V}, \mathcal{E})$ associated with a nodetype mapping $\phi: \mathcal{V} \rightarrow \mathcal{M}$ and an edgetype mapping $\psi: \mathcal{E} \rightarrow \mathcal{R}$. $\mathcal{M}$ and $\mathcal{R}$ denote the predefined sets of nodetypes and edgetypes, respectively, with $|\mathcal{M}| + |\mathcal{R}| > 2$. When $|\mathcal{M}| = 1$ and $|\mathcal{R}| = 1$, it is a homogeneous graph.

\subsection{ Relational graph convolutional network (R-GCN).}

R-GCN ~\citep{schlichtkrull2018modeling} is the ﬁrst method which shows that the GCN framework can be extended to modeling relational data. It is proposed to solve the task of entity classification (assigning types or categorical properties to entities) and link prediction (recovery of the missing triples) on the knowledge graphs. Obviously, this method can be easily transferred to solve the task of node classification and link prediction on the heterogeneous graph. 

In the R-GCN model, the author proposed to use the following forward propagation model to update the status of the nodes in the multi-relation graph.
\begin{equation}
	\label{eq:rgcn}
	h_t'=\sigma(\sum\limits_{r\in\mathcal{R}} \sum_{s_i^r\in \mathcal{N}_t^r } \frac{1}{\lvert \mathcal{N}_t^r \rvert} W_{rel}^{r} h_{s_i^r} + W_{node}h_t),
\end{equation}
where $t$ is the target node, $\mathcal{N}_t^r$ denotes the set of neighbors of node $t$ under the relation $r\in\mathcal{R}$, while $\mathcal{R}$ is the set of all the relations. $s_i^r$ represents the i-th neighbor node of the target node $t$ under the relation $r$. $W_{rel}^{r}$ is the relation-specific parameter matrix and $W_{node}$ is the node transform parameter matrix. $\sigma$ is an element-wise activation function, such as $\textrm{ReLU}(\cdot)$.
Intuitively, equation (\ref{eq:rgcn}) shows how the status $h_t$ of the node $t$ is abtained by aggregating the status of the neighbor nodes for all the relations and then updates to the new status $h_t'$.

It should be pointed out that we can regard R-GCN as a migration of spatial average aggregation GNN(like \cite{kipf2016semi}, \cite{hamilton2017inductive}) on heterogeneous graph. When $\lvert \mathcal{R} \rvert=1$, equation (\ref{eq:rgcn}) becomes the following naive GNN.
\begin{equation}
	\label{eq:rgcn-simple}
	h_t'=\sigma( \frac{1}{\lvert \mathcal{N}_t \rvert} \sum_{s_i\in \mathcal{N}_t } Wh_{s_i} + W_{node}h_t),
\end{equation}

\section{Proposed Method : R-GSN}
\begin{figure*}[ht]
	\centering
	\graphicspath{{figs/}}
	\includegraphics[scale=0.198]{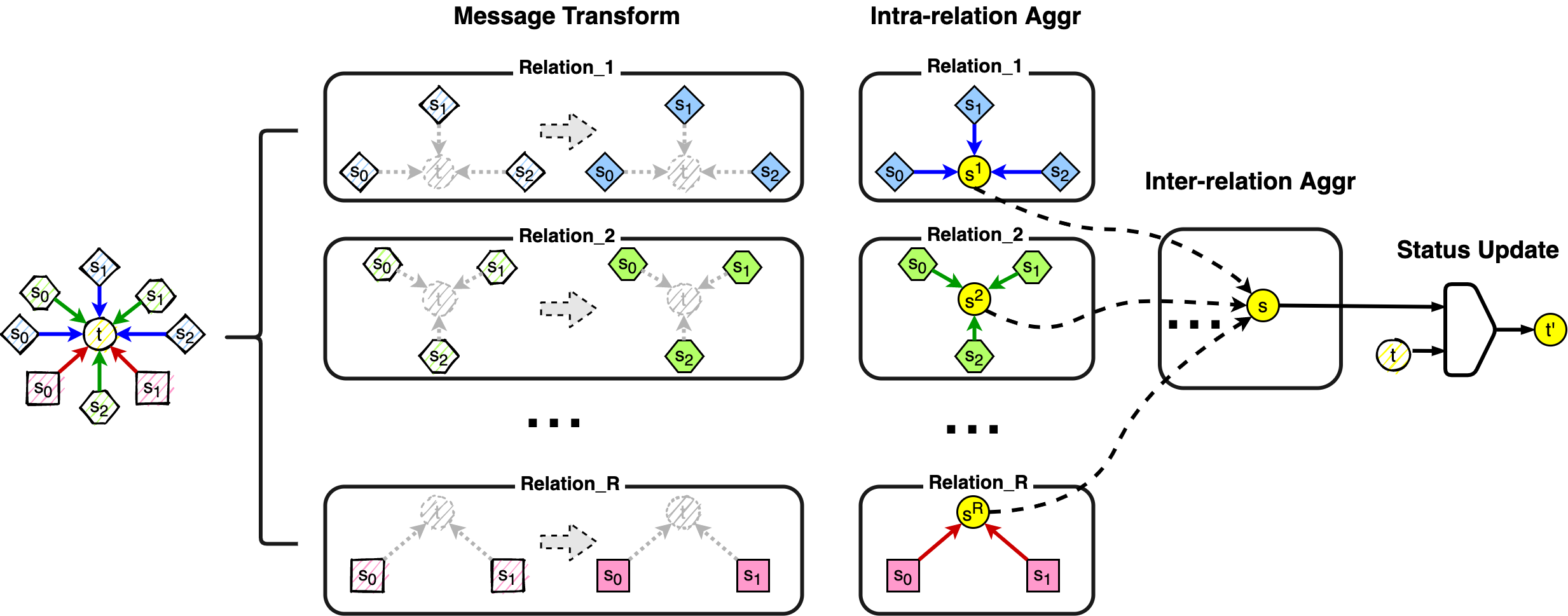}
	\caption{General heterogeneous message passing paradigm. It is summarized into the following four steps. Message Transform, Intra-relation Aggregation, Inter-relation Aggregation and Status Update.}
	\label{fig:hmpgp}
\end{figure*}
\subsection  { General heterogeneous message passing paradigm}

 Before introducing our R-GSN algorithm, in this part we first define a general paradigm for heterogeneous graph neural network message passing(see Figure \ref{fig:hmpgp}). Both R-GCN and our R-GSN are special instances under this paradigm. Heterogeneous message passing is summarized into the following four steps. {\bf Message Transform , Intra-relation Aggregation, Inter-relation Aggregation and Status Update.} Now we will take R-GCN as an example to analyze the detailed operation of each component under this paradigm.

(1) Message Transform 


Message transform is the first step in each layer of heterogeneous message passing. For R-GCN, it is to assign the relation-specific parameter matrix $W_{rel}^r$ to each relation $r$ in the relation set $\mathcal{R}$. for the target node $t$, all neighbor nodes under any relation $r$ will be multiplied by the relation-specific parameter matrix $W_{rel}^r$(\ref{eq:rgcn-1}).
\begin{equation}
	\label{eq:rgcn-1}
	\widetilde{h_{s_i^r}}=W_{rel}^{r} h_{s_i^r}
\end{equation}
the relation-specific parameter matrix can transform the dimension of the feature(status). when 
$h_{s_i^r} \in \mathcal{\R}^d$ and $W_{rel}^r \in \mathcal{\R}^{d'\times d}$, we can get
$\widetilde{h_{s_i^r}} \in \R^{d'}$. In this way, the generated message feature to be aggregated are obtained.

(2) Intra-relation Aggregation

In heterogeneous graph, for the target node $t$, there may be multiple types of edges connected to $t$, and we will separately aggregate the neighbor nodes under each relation $r$. The core of intra-relation aggregation is how to perform feature aggregation for a certain type of edge. For R-GCN, we can treat its intra-relation aggregation as {\bf MEAN aggregation}(\ref{eq:rgcn-2}).
\begin{equation}
	 \label{eq:rgcn-2}
	 h_s^r = \frac{1}{\lvert \mathcal{N}_t^r \rvert} \sum_{s_i^r\in \mathcal{N}_t^r }\widetilde{h_{s_i^r}}
\end{equation}
where $\widetilde{h_{s_i^r}}$ is the transformed feature generated by (\ref{eq:rgcn-1}), 
$\mathcal{N}_t^r$ is the set of the neighbor nodes of target node $t$ under the relation $r$. 
the result of Mean aggregation is $h_s^r$, which represents the neighborhood aggregation feature of the target node $t$ under the relation $r$.

(3) Inter-relation Aggregation 

When the intra-relation aggregation is finished, we will get the aggregation feature $h_s^r$ of the target node $t$ under each relation. After that, we need to aggregate the features obtained under different relation again, which is the core operation of inter-relation.
For R-GCN, its inter-relation aggregation is {\bf SUM aggregation}(\ref{eq:rgcn-3}).
\begin{equation}
 \label{eq:rgcn-3}
h_s = \sum\limits_{r\in\mathcal{R}}h_s^r
\end{equation}
Intuitively, the aggregated features under different relations are aggregated again through the addition operation, and the final aggregated feature $h_s$ of all neighboring nodes of the target node $t$ is obtained.

(4) Status Update

Status update is the last step. We need to merge the feature $h_t$ of the target node $t$ with the feature $h_s$ aggregated from all neighboring nodes together to obtain the final updated feature of the target node $t$, which is the output status $h_t'$(\ref{eq:rgcn-4}).
\begin{equation}
 \label{eq:rgcn-4}
h_t' = \sigma(h_s + W_{node}h_t)
\end{equation}
where $W_{node}$ is the node transform parameter matrix. $\sigma$ is an element-wise activation function, such as $\textrm{ReLU}(\cdot)$.

We find that equation (\ref{eq:rgcn}) is equivalent to concatenating equation (\ref{eq:rgcn-1})(\ref{eq:rgcn-2})(\ref{eq:rgcn-3}) and (\ref{eq:rgcn-4}). We can easily decompose and analyze R-GCN according to the general paradigm defined above.
\graphicspath{{figs/}}

\subsection{Relation-based graph similar network(R-GSN)}
\label{sec:rgsn}
The R-GSN we proposed is a structure that conforms to the general heterogeneous message passing paradigm. it is based on R-GCN, further introduces attention mechanism and similarity aggregation, as well as various normalizations. Next, we will continue to decompose the R-GSN algorithm according to the general paradigm.

(1) Message Transform

The message transformation of R-GSN is the same as that of R-GCN. it is to assign the relation-specific parameter matrix $W_{rel}^r$ to each relation $r$ in the relation set $\mathcal{R}$. for the target node $t$, all neighbor nodes under any relation $r$ will be multiplied by the relation-specific parameter matrix $W_{rel}^r$(\ref{eq:rgsn-1}).
\begin{equation}
\label{eq:rgsn-1}
\widetilde{h_{s_i^r}}=W_{rel}^{r} h_{s_i^r}
\end{equation}
where $h_{s_i^r} \in \mathcal{\R}^d$ and $W_{rel}^r \in \mathcal{\R}^{d'\times d}$, so
$\widetilde{h_{s_i^r}} \in \R^{d'}$. In this way, the generated message feature to be aggregated are obtained.

(2) Intra-relation Aggregation

\begin{figure*}[h]
\centering
\graphicspath{{figs/}}
\includegraphics[scale=0.22]{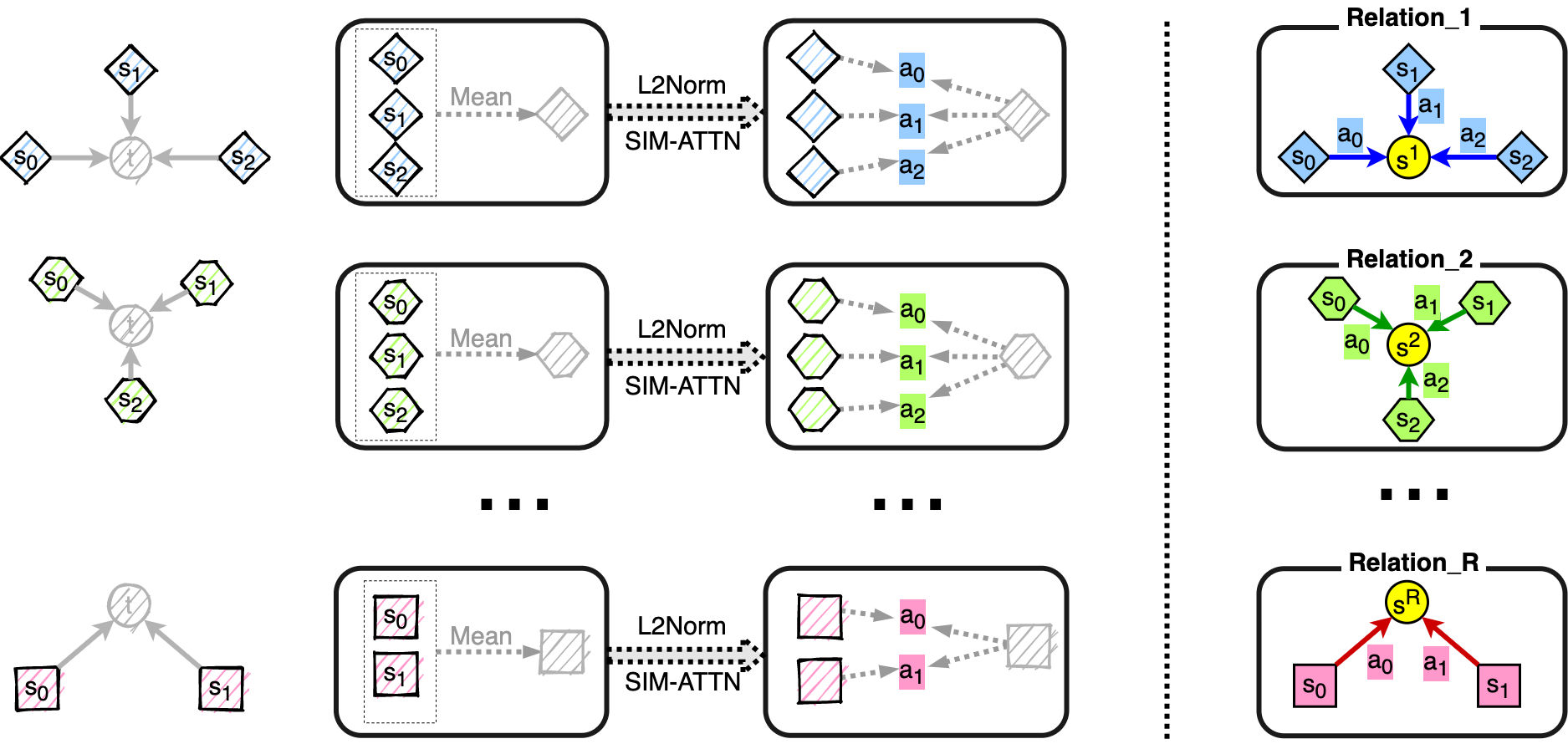}
\caption{R-GSN Intra-realtion Aggregation(SIM-ATTN aggregation). }
\label{fig:rgsn-intra}
\end{figure*}
The intra-relation aggregation of R-GCN is MEAN aggregation described in equation (\ref{eq:rgcn-2}). In R-GSN, the intra-relation aggregation is {\bf SIM-ATTN aggregation}(similarity-attention aggregation), which can also be regarded as a coefficient-weighted aggregation. As described in Figure \ref{fig:rgsn-intra} and equation (\ref{eq:rgsn-2}).
\begin{equation}
\label{eq:rgsn-2}
h_s^r = \mathrm{L2Norm}( \sum_{s_i^r\in \mathcal{N}_t^r } a_i^r  \widetilde{h_{s_i^r}})
\end{equation}
where $\mathrm{L2Norm}$ is L2 normalization($\mathrm{L2Norm}(\vx) = \tfrac{\vx}{||\vx||_2}$). $\widetilde{h_{s_i^r}}$ is the transformed feature generated by equation (\ref{eq:rgsn-1}). 
the result of SIM-ATTN aggregation is $h_s^r$, which represents the neighborhood aggregation feature of the target node $t$ under the relation $r$, and $h_s^r \in \R^{d'}$. The difference from R-GCN's MEAN aggregation is that there is an additional weighted coefficient $a_i^r$, which can be calculated by the following equation (\ref{eq:rgsn-2-1})(\ref{eq:rgsn-2-2})(\ref{eq:rgsn-2-3})(\ref{eq:rgsn-2-4}).
\begin{gather}
\label{eq:rgsn-2-1}
\ddot{h_{s_i^r}} = \mathrm{L2Norm}(h_{s_i^r}) 
\\
\label{eq:rgsn-2-2}
\ddot{h_{s^r}} = \mathrm{L2Norm}( \frac{1}{\lvert \mathcal{N}_t^r \rvert} \sum_{s_i^r\in \mathcal{N}_t^r } h_{s_i^r})
\\
\label{eq:rgsn-2-3}
e_i^r = attn^r*[\ddot{h_{s_i^r}} || \ddot{h_{s^r}}]
\\
\label{eq:rgsn-2-4}
a_i^r = \mathrm{softmax}(e_i^r) = \frac{e_i^r}{\sum_{j\in \mathcal{N}_t^r}{e_j^r}}
\end{gather}
where $h_{s_i^r} \in \R^d$, so $\ddot{h_{s_i^r}} \in \R^d$ and $\ddot{h_{s^r}} \in \R^d$, symbol $||$ means concatenation, so $[\ddot{h_{s_i^r}} || \ddot{h_{s^r}}] \in \mathcal{\R}^{2d}$, $attn^r \in \R^{2d}$ is the trainable attention mechanism parameters which is also relation-specific, the $*$ here represents the inner product. When $e_i^r$ is calculated, the final weighted coefficient $a_i^r$ can be obtained through the neighborhood softmax on $e_i^r$. 

Note that our SIM-ATTN aggregation method is a little similar to GAT's attention aggregation(\ref{eq:gat-1})(\ref{eq:gat-2}) ~\citep{velivckovic2017graph}, but there is still a big difference. For example, in the process of calculating the weighted coefficient, only the feature of the source node is used, and the feature of the target node is not used. Before similarity calculation, we first performed the L2 normalization operation in order to make the features concatnate at the same scale. Meanwhile in SIM-ATTN aggregation, the Leakey ReLU operation is abandoned.
\begin{gather}
\label{eq:gat-1}
e_i =\mathrm{LeakyReLU}(attn*[W {h_{s_i}} || W h_t])
\\
\label{eq:gat-2}
a_i = \mathrm{softmax}(e_i) = \frac{e_i^r}{\sum_{j\in \mathcal{N}_t^r}{e_j^r}}
\end{gather}

(3) Inter-relation Aggregation 

\begin{figure*}[h]
\centering
\graphicspath{{figs/}}
\includegraphics[scale=0.2]{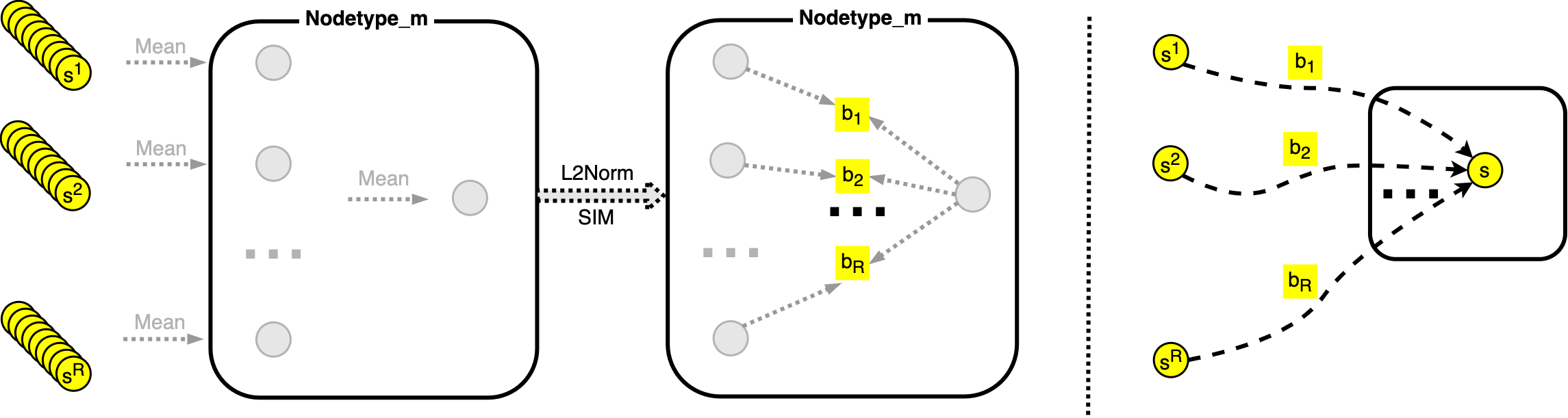}
\caption{R-GSN Inter-realtion Aggregation(SIM aggregation). }
\label{fig:rgsn-inter}
\end{figure*}
As mentioned above, inter-ralation aggregation will aggregate the feature $h_s^r$ of the target node t under all relations. In R-GCN, SUM aggregation is used. In R-GSN, we propose to use {\bf SIM aggregation}, which is somewhat similar to SIM-ATTN aggregation. The only difference is that no attention vector is introduced here, so there are no parameters. We can use equation (\ref{eq:rgsn-3}) to describe SIM aggregation (Figure \ref{fig:rgsn-inter}).
\begin{equation}
\label{eq:rgsn-3}
h_s = \sum\limits_{r\in\mathcal{R}} b_r^m  h_s^r
\end{equation}
 Here $h_s \in \R^{d'}$. The difference between R-GSN(\ref{eq:rgsn-3}) and R-GCN(\ref{eq:rgcn-3})'s inter-relation aggregation is $b_r^m$. 
$b_r^m$ is also a similarity weighted coefficient, which represents the importance weight that the nodetype $m$ should be given under the relation $r$. Of course, the premise is that assuming the nodetype of the target node $t$ is $m$. 
$m \in \mathcal{M}$ and  $\mathcal{M}$ is a collection of node types. $b_r^m$ can be calculated by the following equation (\ref{eq:rgsn-3-1})(\ref{eq:rgsn-3-2})(\ref{eq:rgsn-3-3})(\ref{eq:rgsn-3-4}).
\begin{gather}
\label{eq:rgsn-3-1}
\ddot{h_{s^r}^m} = \mathrm{L2Norm}(\frac{1}{|\mathcal{V}^m|} \sum_{t_j \in \mathcal{V}^m} h_{s^r}(t_j)) 
\\
\label{eq:rgsn-3-2}
\ddot{h_{s}^m} = \mathrm{L2Norm}(\frac{1}{|\mathcal{R}|\times|\mathcal{V}^m|} \sum_{r \in \mathcal{R}} \sum_{t_j \in \mathcal{V}^m} h_{s^r}(t_j)) 
\\
\label{eq:rgsn-3-3}
e_r^m = \ddot{h_{s^r}^m} * \ddot{h_{s}^m}
\\
\label{eq:rgsn-3-4}
b_r^m = \mathrm{softmax}(e_r^m) = \frac{e_r^m}{\sum_{j\in \mathcal{R}}{e_j^m}}
\end{gather}
We mentioned that $h_s^r$ is the neighborhood aggregated feature of the target node $t$ under the relation $r$, We can also write $h_s^r$ as a verbose form $h_s^r(t)$. In equation (\ref{eq:rgsn-3-1}), $\mathcal{V}^m$ represents the collection of all nodes of nodetype $m$. Where $h_{s^r}(t_j) \in \R^d$, so $\ddot{h_{s^r}^m} \in \R^d$ and $\ddot{h_{s}^m} \in \R^d$, the $*$ here represents the inner product. When $e_r^m$ is calculated, the final weighted coefficient $b_r^m$ can be obtained through the softmax on $e_r^m$ across all relations. 

(4) Status Update

\begin{figure}[h]
\centering
\graphicspath{{figs/}}
\includegraphics[scale=0.33]{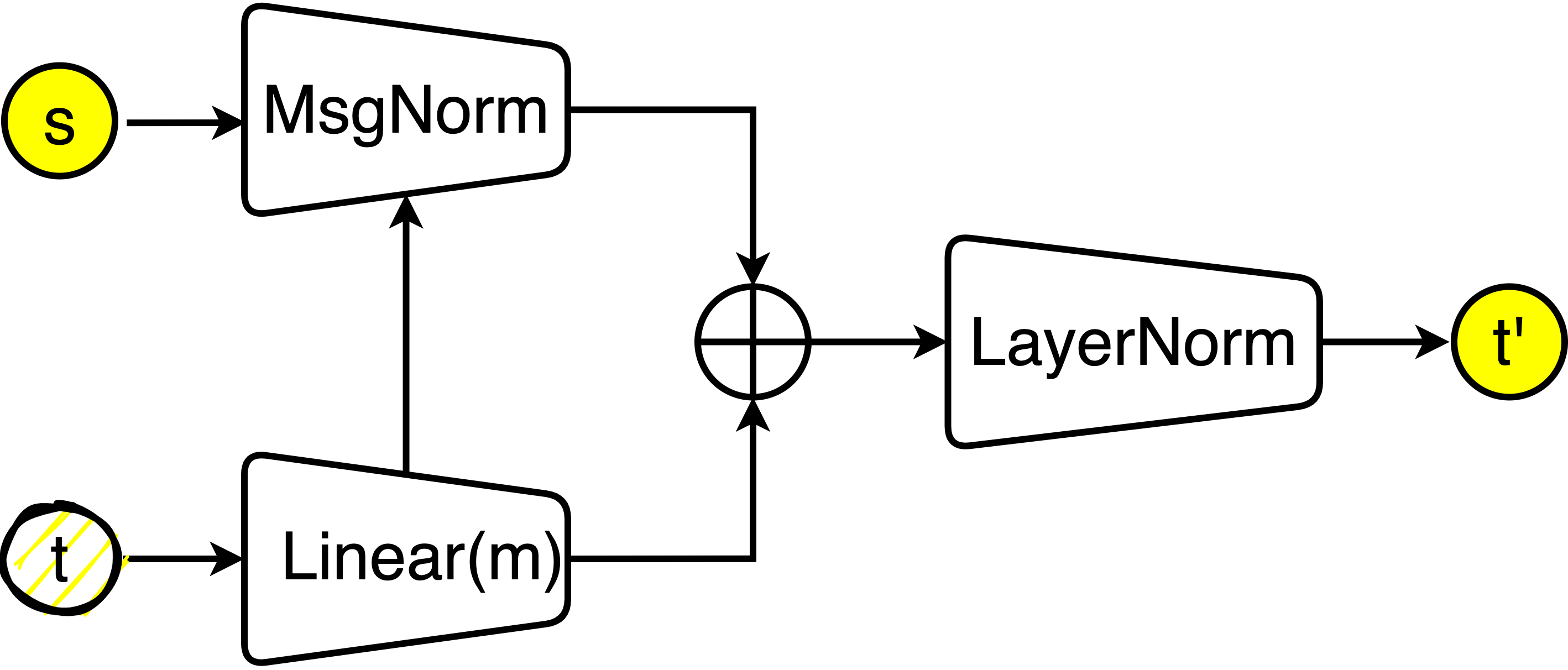}
\caption{R-GSN Status Update. }
\label{fig:rgsn-update}
\end{figure}
For R-GSN, the process of State Update is described in equation (\ref{eq:rgsn-4}) and Figure \ref{fig:rgsn-update}.
\begin{equation}{}
	 \label{eq:rgsn-4}
	h_t' = \sigma(\mathrm{LayerNorm}(\mathrm{MsgNorm}(h_s, W_{node}^m h_t)  + W_{node}^m h_t))
\end{equation}
where $W_{node}^m$ is nodetype-specific parameter matrix for target node $t$ of nodetype $m$, $W_{node}^m \in \R^{d' \times d}$. MsgNorm is proposed by \cite{li2020deepergcn}, its core idea is to normalize the message feature first, and then multiply it by the norm of the target node feature. In R-GSN, we design a normalization layer that cascades MsgNorm and LayerNorm ~\citep{ba2016layer}, which can integrate message features and node features well. $\sigma$ is an element-wise activation function, such as $\textrm{ReLU}(\cdot)$. $h_t'$ is the updated feature of the target node $t$ and $h_t' \in \R^{d'}$.

\section{Experments}

\subsection{Heterogeneous graph dataset: ogbn-mag}
\begin{figure}[h]
\centering
\graphicspath{{figs/}}
\includegraphics[scale=0.4]{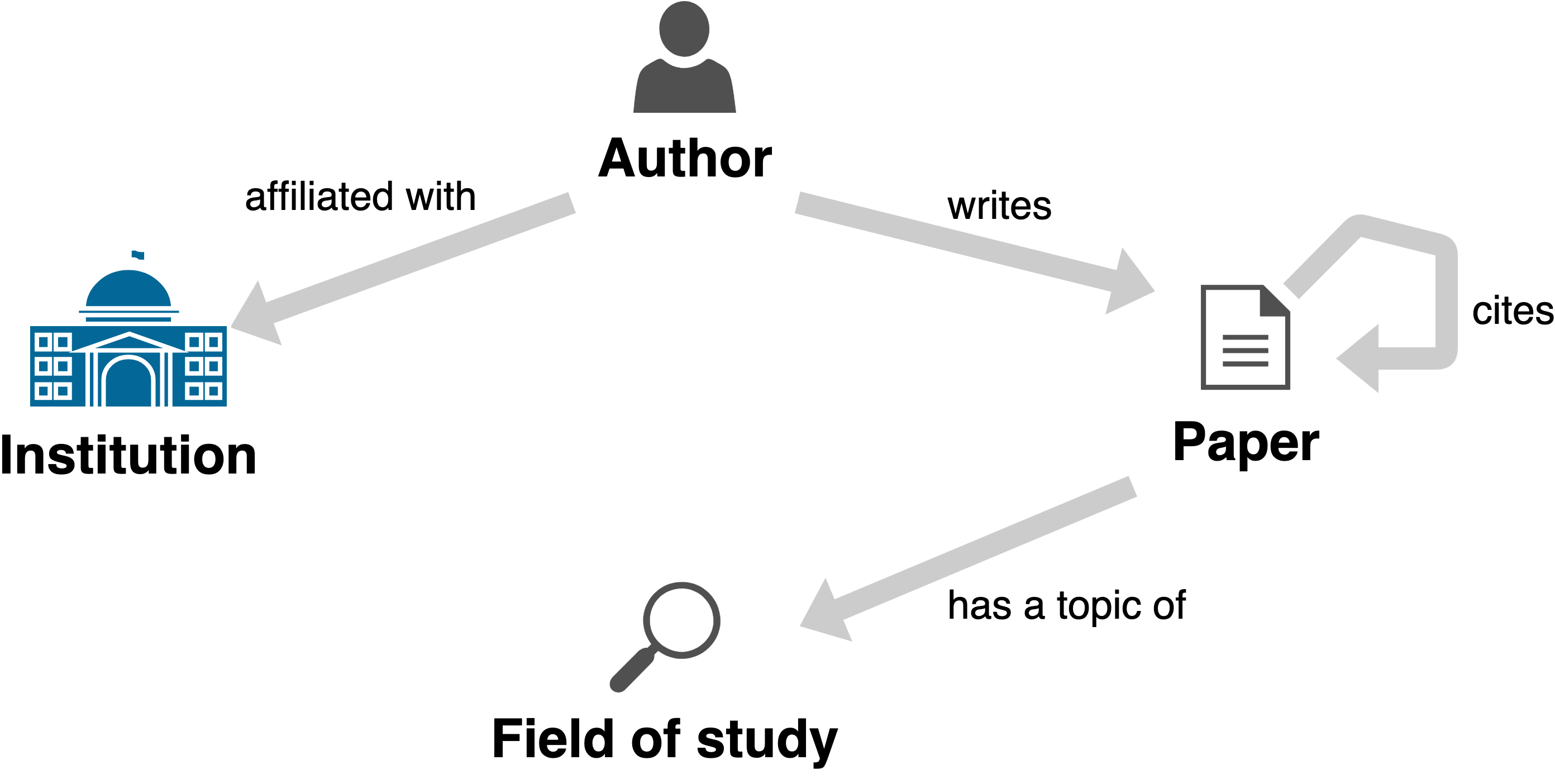}
\caption{The schema of $\texttt{ogbn-mag}$  dataset. }
\label{fig:mag}
\end{figure}
The $\texttt{ogbn-mag}$ dataset is a heterogeneous network composed of a subset of the Microsoft Academic Graph (MAG) ~\citep{wang2020microsoft}. It contains four types of entities—papers (736,389 nodes), authors (1,134,649 nodes), institutions (8,740 nodes), and ﬁelds of study (59,965 nodes)—as well as four types of directed relations connecting two types of entities—an author “is afﬁliated with” an institution, an author “writes” a paper, a paper “cites” a paper, and a paper “has a topic of” a ﬁeld of study. each paper is associated with a 128-dimensional word2vec feature vector, and all the other types of entities are not associated with input node features.The schema of the $\texttt{ogbn-mag}$ dataset is shown in Figure \ref{fig:mag}, and the statistical information is shown in Table \ref{tb:mag-node-stat} and \ref{tb:mag-edge-stat}.
\vskip -0.1in
\begin{table}[htbp]
	\footnotesize
	\centering
	\caption{$\texttt{ogbn-mag}$ node statistics}
	\label{tb:mag-node-stat}
		\vskip 0.1in
	    \setlength{\tabcolsep}{2.1mm}{
		\begin{tabular}{lrr}
		\toprule
		\textbf{Node type(4)} & \textbf{Number} & \textbf{description}                                                          \\ \midrule
		Paper                 & 736389          & 128-d feature, 349 categories \\
		Author                & 1134649         & None                                                                          \\
		Institution           & 8740            & None                                                                          \\
		Field   of study      & 59965           & None                                                                          \\ \bottomrule
		\end{tabular}}	
\end{table}
\vskip -0.1in
\begin{table}[htbp]
	\footnotesize	
	\centering
	\caption{$\texttt{ogbn-mag}$ edge statistics}
	\label{tb:mag-edge-stat}
		\vskip 0.1in
	    \setlength{\tabcolsep}{4mm}{	
		\begin{tabular}{lr}
		\toprule
		\textbf{Edge type(4)} & \textbf{Number}    \\ 
		\midrule
		'Author', 'affiliated with', 'Institution'  & 1043998 \\
		'Author', 'writes', 'Paper'                 & 7145660 \\
		'Paper', 'cites', 'Paper'                   & 5416271 \\
		'Paper', 'has a topic of', 'Field of study' & 7505078 \\
		\bottomrule
		\end{tabular}}
\end{table}

The task on the heterogeneous $\texttt{ogbn-mag}$ dataset is to predict the venue (conference or journal) of each paper, given its content, references, authors, and authors’ afﬁliations. This is of practical interest as some manuscripts’ venue information is unknown or missing in MAG, due to the noisy nature of Web data. In total, there are 349 different venues in $\texttt{ogbn-mag}$, making the task a 349-class classiﬁcation problem.

\subsection{Node classification }

\subsubsection{R-GSN implementation details}
In R-GSN, the implementation details of each heterogeneous message passing layer have been described in section \ref{sec:rgsn}. Our R-GSN model has two layers, the feature dimension of the hidden layer is 64, and the feature dimension of the output layer is 349 (the number of categories of the $\texttt{ogbn-mag}$ dataset). We use cross entropy loss and Adamw~\citep{loshchilov2018fixing} optimizer for training, the learning rate is set to 0.004, and use early-stop technology and Dropout~\citep{srivastava2014dropout} technology to prevent overfitting. We use the minibatch neighborhood sampling method for training~\citep{hamilton2017inductive}, the batchsize is 1024. For the original feature, we set the feature to be trainable and use FLAG~\citep{kong2020flag} adversarial training. The division of train/val/test is in accordance with the official division. Table \ref{tb:ogbn-mag-baseline} reports the accuracy of the validation and test dataset of our R-GSN on the $\texttt{ogbn-mag}$(Note that the report is the mean and standard deviation of the results of 10 random runs ), the results of other algorithms are all from the public results on the official leaderboard (\url{https://ogb.stanford.edu/docs/leader_nodeprop/}).
\vskip -0.1in
\begin{table}[htbp]
	\footnotesize
    \centering
    \caption{ Results for $\texttt{ogbn-mag}$.}
    \label{tb:ogbn-mag-baseline}
    \vskip 0.1in
    \setlength{\tabcolsep}{2mm}{	
    \renewcommand{\arraystretch}{1.2}
    \begin{tabular}{lrcc}
      \toprule
        \mr{2}{\textbf{Method}} & \mc{3}{c}{\textbf{Accuracy (\%)}} \\
         &Params & Validation & \textbf{Test} \\
      \midrule
        \textsc{MLP}         &188,509  & 26.26\std{0.16} & 26.92\std{0.26} \\
        \textsc{GCN}        &1,495,901 & 29.53\std{0.22} & 30.43\std{0.25} \\
        \textsc{GraphSAGE}        &1,495,901 & 30.70\std{0.19} & 31.53\std{0.15} \\
      \midrule
        \textsc{MetaPath2Vec}  &94,479,069 & 35.06\std{0.17} & 35.44\std{0.36} \\
        \textsc{ClusterGCN}  &154,366,772 & 38.40\std{0.31} & 37.32\std{0.37} \\
        \textsc{SIGN}  &3,724,645 & 40.68\std{0.10} & 40.46\std{0.12} \\
        \textsc{R-GCN}  &154,366,772	 & 47.61\std{0.68} & 46.78\std{0.67} \\	
        \textsc{GraphSAINT} &154,366,772 & 48.37\std{0.26} & 47.51\std{0.22} \\
        \textsc{HGT}  &21,173,389	 & 49.89\std{0.47} & 49.27\std{0.61} \\
        \midrule
        {\bf \textsc{R-GSN}}  &154,373,028	 & \textbf{51.82}\std{0.41} & \textbf{50.32}\std{0.37} \\		
      \bottomrule
    \end{tabular}}
\end{table}

Because $\texttt{ogbn-mag}$ is a heterogeneous graph, some changes need to be made to migrate the GNN algorithm on a homogeneous graph. Specifically, for GCN~\citep{kipf2016semi} and GraphSAGE~\citep{hamilton2017inductive}, since they were originally designed for homogeneous graphs, apply the model to homogeneous subgraphs, keeping only the paper nodes and their citation relations. For MLP, the graph structure is ignored and only paper nodes are considered. Except for these three algorithms, the others are directly designed for heterogeneous graphs.

Obviously, R-GSN has a powerful improvement over R-GCN~\citep{schlichtkrull2018modeling}, and its performance is better than the previous best algorithm HGT~\citep{hu2020heterogeneous}. R-GSN achieves the state-of-the-art performance on the $\texttt{ogbn-mag}$ dataset. The average accuracy on the validation dataset and test dataset are respectivly 51.82\% and 50.32\%, both are higher than those of the existing algorithms. At the same time, compared with the baseline R-GCN, there is a 4.21\% gain on the validation dataset and a 3.54\% gain on the test dataset. Compared with HGT, there is a 1.93\% gain on the validation dataset and a 1.05\% gain on the test dataset.

\subsection{Ablation study}
In this part, we show ablation experiments to analyze which operations boost the performance of R-GCN and R-GSN,
The results of ablation experiments are shown in Table \ref{tb:ogbn-mag-ablation}. \textbf{SIM-ATTN}: the method used by R-GSN Intra-relation aggregation. \textbf{SIM}: the method used by R-GSN Inter-relation aggregation. \textbf{Norm}: Normalization is used in 4 places, namely LayerNorm at the input, L2Norm at the Intra-relation aggregation, MsgNorm and LayerNorm at Status Update. \textbf{FT}: the abbreviation of feature training, because the original heterogeneous graph data only has the features of the paper node, so we first perform feature pre-propagation according to the connection relation to obtain the initial features of other types of nodes, and set these features as trainable parameter. \textbf{FLAG}: It is an adversarial training method that improves the robustness of the trained model by perturbing the original data.

\begin{table*}[htbp]
	\footnotesize
    \centering
    \caption{ Ablation study of R-GSN on $\texttt{ogbn-mag}$.}
    \label{tb:ogbn-mag-ablation}
	\vskip 0.1in	
    \setlength{\tabcolsep}{2mm}{
    \renewcommand{\arraystretch}{1.2}
    \begin{tabular}{lcccccccc}
      \toprule
        \mr{2}{\textbf{Method}}	 &\mr{2}{\textbf{SIM-ATTN}} &\mr{2}{\textbf{SIM}} &\mr{2}{\textbf{Norm}} &\mr{2}{\textbf{FT}}	&\mr{2}{\textbf{FLAG}} & \mc{3}{c}{\textbf{Accuracy (\%)}} \\
		& & & & & &Params & Validation & \textbf{Test} \\
      \midrule
      	\textsc{R-GCN} &-- &-- &-- &-- &--              &154,366,772   &47.61\std{0.68}   &46.78\std{0.67}  \\
      \midrule	
      	\textsc{R-GCN(1+)} &-- &-- &\duihao &-- &--          &154,370,340    &49.86\std{0.29} &48.83\std{0.50}   \\
      	\textsc{R-GCN(2+)} &-- &-- &\duihao &\duihao &--   &154,370,340    &50.72\std{0.27}  &49.66\std{0.37}  \\
	    \textsc{R-GCN(3+)} &-- &-- &\duihao  &\duihao &\duihao    &154,370,340     &51.04\std{0.48}  &49.74\std{0.47}      \\	

	  \midrule
        \textsc{R-GSN(1+)}  &\duihao &\duihao &\duihao &-- &--      &154,373,028  & 50.43\std{0.46} & 49.26\std{0.40} \\
        \textsc{R-GSN(2+)}   &\duihao &\duihao &\duihao  &\duihao &--  &154,373,028 & 51.33\std{0.35} & 50.10\std{0.42} \\
        {\bf \textsc{R-GSN(3+)}} &\duihao &\duihao &\duihao &\duihao &\duihao  &154,373,028	 & \textbf{51.82}\std{0.41} & \textbf{50.32}\std{0.37} \\			
      \bottomrule
    \end{tabular}}
\end{table*}

\begin{figure*}[h]
\centering
\graphicspath{{figs/}}
\includegraphics[scale=0.45]{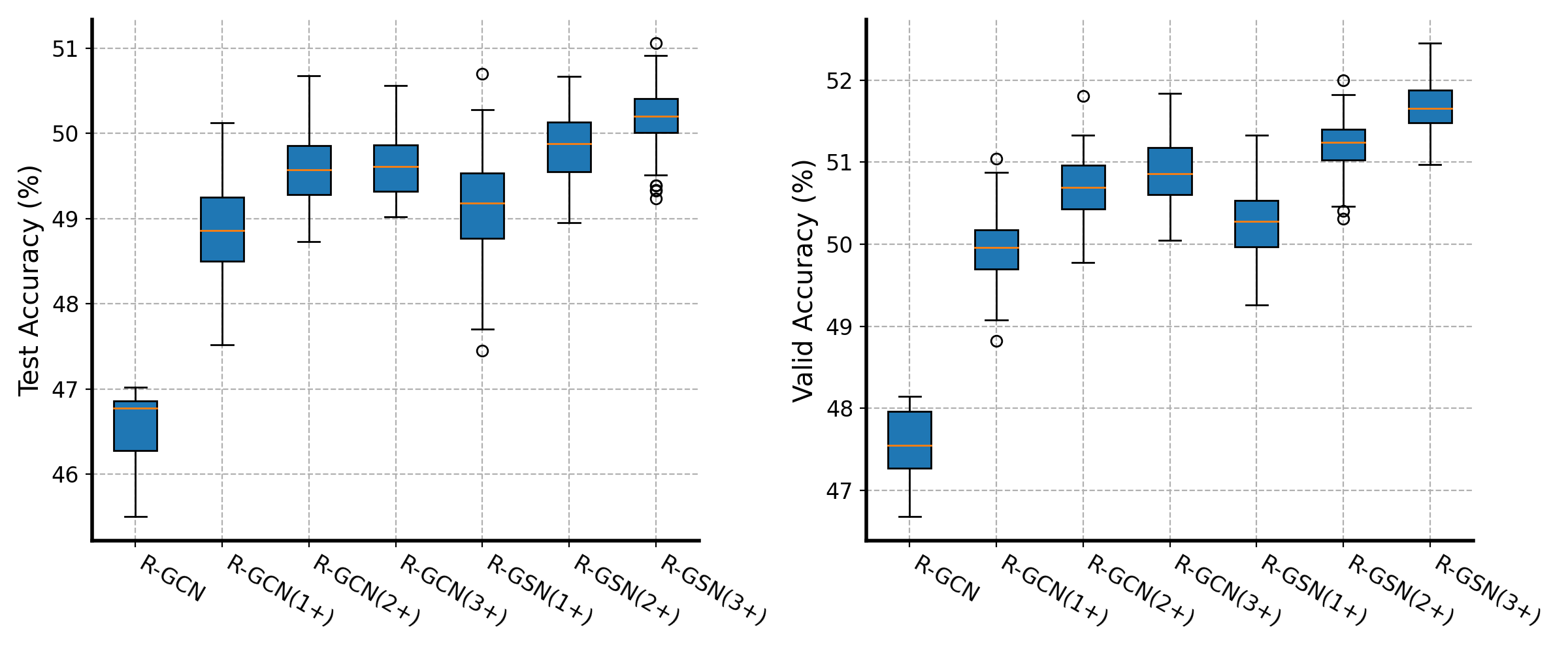}
\caption{Ablation study. Left: Test accuracy on the $\texttt{ogbn-mag}$ dataset. Right: Valid accuracy on the $\texttt{ogbn-mag}$ dataset. }
\label{fig:ablation}
\end{figure*}

In Table \ref{tb:ogbn-mag-ablation}, the result of R-GCN is the result reported on the official website, and we use it as the baseline model. R-GCN(1+), R-GCN(2+) and R-GCN(3+) are R-GCN with one, two and three operations added, namely Norm, FT, FLAG . We can find that when three operations are successively added to R-GCN and R-GSN, the accuracy of the validation and test dataset will increase accordingly. Therefore, these three operations all have the positive effect. Of course, by comparing R-GCN(1+) and R-GSN(1+), R-GCN(2+) and R-GSN(2+), R-GCN(3+) and R-GSN(3+), it can be found that R-GSN with SIM-ATTN aggregation and SIM aggregation has better performance than R-GCN with MEAN aggregation and SUM aggregation. Figure \ref{fig:ablation} shows more intuitively in Table \ref{tb:ogbn-mag-ablation}. The boxplot of the accuracy distribution of the 7 models after multiple random trainings (more than 10 times).

\subsection{Homogeneous graph version of R-GSN: GSN}
We hope that the aggregation method proposed in R-GSN and some tricks should also be beneficial to homogeneous graphs. To this end, we conducted experiments on the paper citation network: $\texttt{ogbn-arxiv}$ dataset, which is significantly larger than the Cora, CiteSeer, and PubMed datasets, and is more in line with real-world applications. Small datasets such as Cora often have only hundreds of thousands of edges, while the $\texttt{ogbn-arxiv}$ has 169,343 nodes and 1,162,243 edges, with an average degree of 13.7.

\begin{figure*}[htbp]
\centering
\graphicspath{{figs/}}
\includegraphics[scale=0.45]{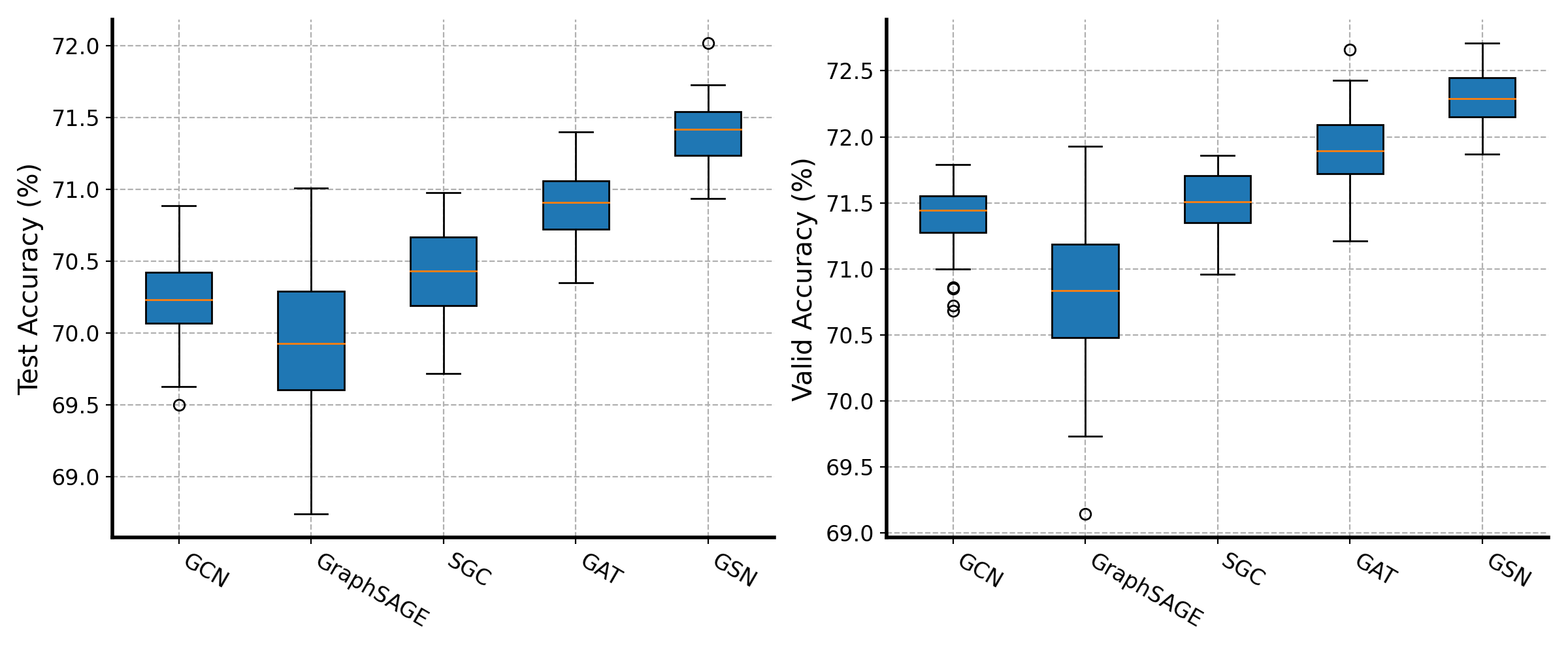}
\caption{Results on the $\texttt{ogbn-arxiv}$ dataset. Left: Test accuracy. Right: Valid accuracy. }
\label{fig:arxiv}
\end{figure*}

The $\texttt{ogbn-arxiv}$ dataset is a directed graph, which is a citation network between ARXIV papers. Each node is a paper published on ARXIV, and each directed edge represents the paper citation relationship between the two papers. Each paper has a 128-dimensional initial feature vector. The Word2vec model is run on the MAG corpus to calculate the embedding of a single word. The initial feature vector is obtained by averaging the embeddings of the words in its title and abstract. The task is to predict 40 subject areas of computer science papers on ARXIV, such as artificial intelligence, operating systems, and etc., so this is a multi-classification task with 40 categories.

We designed a homogenous version of R-GSN, called GSN. It can be regarded as the R-GSN with only one type of node and one type of edge. We conducted a simple experiment on the $\texttt{ogbn-arxiv}$ dataset and compared it with several classic graph neural network algorithms. The experimental results are shown in Table \ref{tb:ogbn-arxiv}, which reported the mean and standard deviation of the accuracy after 10 random trainings. 
\vskip -0.1in
\begin{table}[htbp]
	\footnotesize
    \centering
    \caption{ Results for $\texttt{ogbn-arxiv}$.}
    \label{tb:ogbn-arxiv}
    \vskip 0.1in
    \setlength{\tabcolsep}{3mm}{	
    \renewcommand{\arraystretch}{1.2}
    \begin{tabular}{lrcc}
      \toprule
        \mr{2}{\textbf{Method}} & \mc{3}{c}{\textbf{Accuracy (\%)}} \\
         &Params & Validation & \textbf{Test} \\
      \midrule
        \textsc{MLP}  &38,696 & 54.22\std{0.12} & 56.25\std{0.12} \\
        \textsc{GCN}  &38,184 & 70.24\std{0.28} & 71.39\std{0.22} \\
        \textsc{GraphSAGE}  &76,072	 & 69.90\std{0.49} & 70.81\std{0.50} \\	
        \textsc{SGC} &38,184 & 70.43\std{0.30} & 71.51\std{0.23} \\
        \textsc{GAT}  &38,776	 & 70.89\std{0.26} & 71.89\std{0.27} \\
        \midrule
        {\bf \textsc{GSN}}  &39,547	 & \textbf{71.40}\std{0.22} & \textbf{72.31}\std{0.21} \\		
      \bottomrule
    \end{tabular}}
\end{table}

For a fair comparison, all the graph neural network algorithms in the table \ref{tb:ogbn-arxiv} are unified to 3 layers, the hidden layer dimension is 128, the Adam optimizer is used, the learning rate is 0.01, and the Dropout and Early-stop techniques are adopted to prevent overfitting. It can be found that our GSN has better results than MLP and these classic graph neural network algorithms, which proves the effectiveness of our proposed aggregation method and several tricks. The box plot in Figure \ref{fig:arxiv} can more intuitively show the performance distribution of these graph neural network models after multiple random trainings (more than 10 times). It can be seen that GSN has the best performance on both the validation and the test dataset.

\subsection{Extended general heterogeneous message passing paradigm}
The existing heterogeneous graph neural network al- gorithms mainly have two ideas, one is based on meta-path and the other is not. The idea based on meta-path often requires a lot of manual preprocessing, at the same time it is difficult to extend to large scale graphs. 

The general heterogeneous message passing paradigm and the R-GSN algorithm that we introduced earlier are all based on relation, and meta-path can be seen as a generalized higher-order relation. In this part, we appropriately extend the previous general heterogeneous message passing paradigm, which can be adjusted to the following four steps: Message Transform, Intra-metapath Aggregation, Inter-metapath Aggregation and Status Update. At the same time, we propose the M-GSN algorithm under this framework. In the M-GSN algorithm, two aggregation methods are used in the Intra-metapath Aggregation, MEAN aggregation and SIM aggregation. 

\begin{figure}[htbp]
\centering
\graphicspath{{figs/}}
\includegraphics[scale=0.7]{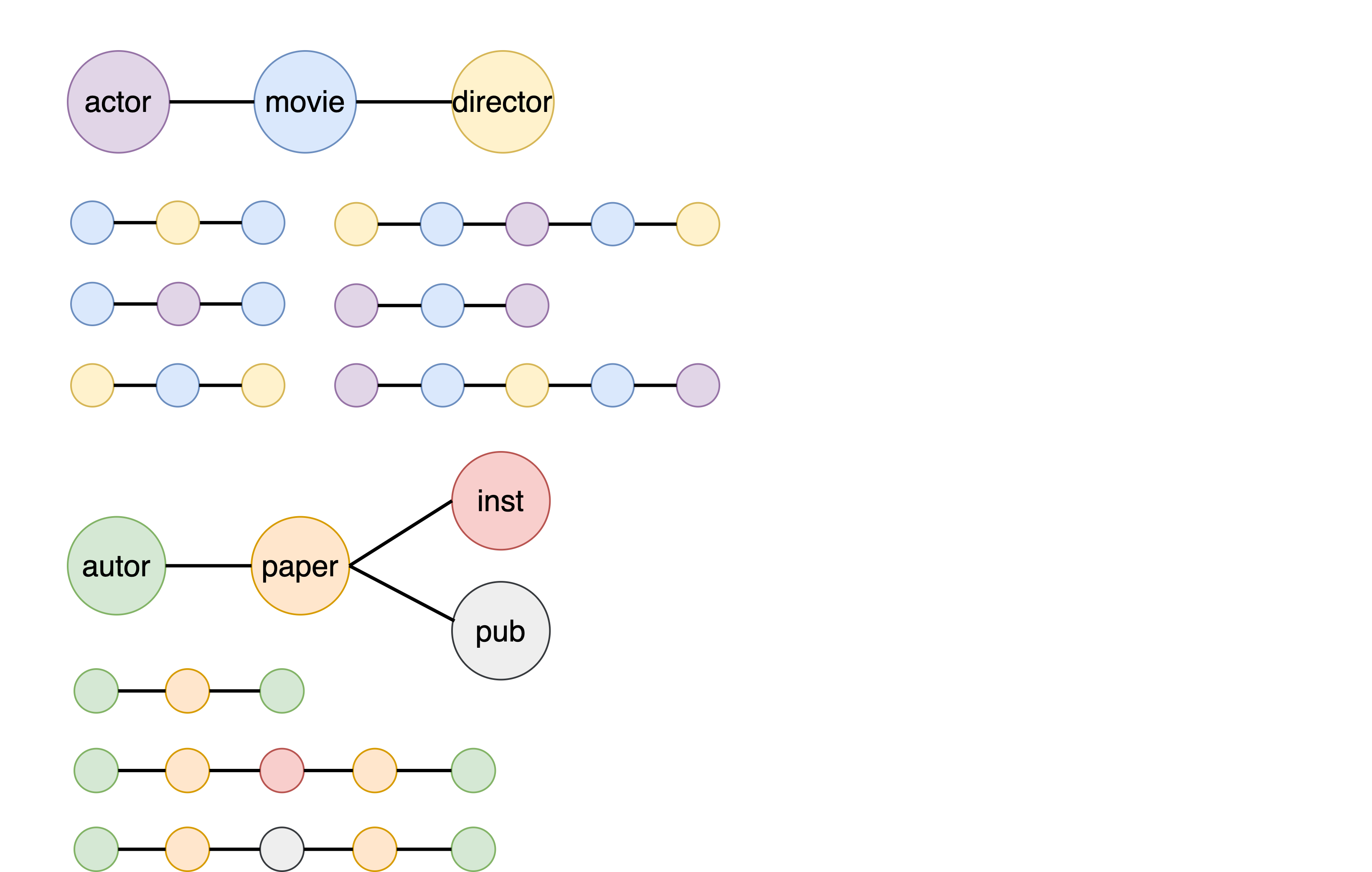}
\caption{The schema and meta-path of $\texttt{IMDB}$ and $\texttt{DBLP}$ datasets}
\label{fig:imdb and dblp}
\end{figure}

We used two datasets for experiments, $\texttt{IMDB}$ and $\texttt{DBLP}$, and the preprocessing of the datasets follows the way in the MAGNN~\citep{yun2019graph}. $\texttt{IMDB}$ is an online dataset about movies and TV shows. We use a subset of $\texttt{IMDB}$ here, which contains 4278 movies, 208 directors, and 5257 actors after data preprocessing. Movies are divided into three categories (action, comedy and drama). Each movie is represented by a bag of words describing its keywords. For the semi-supervised learning model, the movie node is divided into train/val/test sets, which contain 400 (9.35\%), 400 (9.35\%) and 3478 (81.30\%) training nodes, validation nodes and test nodes, respectively. $\texttt{DBLP}$ is a computer science website. We use a subset of $\texttt{DBLP}$. After data preprocessing, it contains 4057 authors, 14328 papers, 7723 institutions and 20 publication locations. The author is divided into four research fields (database, data mining, artificial intelligence, information retrieval). Each author is described by a bag of words of keywords in their papers. For the semi-supervised learning model, the author nodes are divided into 400 (9.35\%), 400 (9.35\%) and 3257 (80.28\%) training nodes, validation nodes and test nodes. The schema and meta-path of $\texttt{IMDB}$ and $\texttt{DBLP}$ datasets are shown in Figure \ref{fig:imdb and dblp}.

We conducted node classification experiments on the $\texttt{IMDB}$ and $\texttt{DBLP}$ datasets. After training, we use the trained model to extract the feature of the nodes in the test set, and then divide the data according to different train/test ratios (20\%, 40\%, 60\%, 80\%) are sent to the linear SVM model for training. It should be noted that in order to make a fair comparison, we just sent the nodes of the test dataset to the SVM model, because our semi-supervised graph embedding algorithm has seen the nodes of the training dataset during training. Therefore, the train/test ratio of the SVM here only involves the test dataset (that is, 3478 nodes for $\texttt{IMDB}$ and 3257 nodes for $\texttt{DBLP}$). We report the average Macro-F1 and Micro-F1 of 10 runs for each embedded model in Table \ref{tb:mgsn}.


\begin{table*}[h]
	\footnotesize
    \centering
    \caption{ Results for $\texttt{IMDB}$ and $\texttt{DBLP}$.}
    \label{tb:mgsn}
    \vskip 0.1in
    \setlength{\tabcolsep}{2.1mm}{	
    \renewcommand{\arraystretch}{1.2}
\begin{tabular}{lccccccccccc}
\toprule
\multicolumn{1}{c}{Datasets} & \multicolumn{1}{c}{Metrics}                                             & \multicolumn{1}{c}{Train \%} & \multicolumn{1}{c}{GCN} & \multicolumn{1}{c}{GAT} & \multicolumn{1}{c}{HAN} & \multicolumn{1}{c}{\begin{tabular}[c]{@{}c@{}}MAGNN\\ paper\end{tabular}} & \multicolumn{1}{c}{\begin{tabular}[c]{@{}c@{}}MAGNN\\ repro\end{tabular}} & \multicolumn{1}{c}{\begin{tabular}[c]{@{}c@{}}\textbf{M-GSN}\\ \textbf{Mean}\end{tabular}} & \multicolumn{1}{c}{\begin{tabular}[c]{@{}c@{}}\textbf{M-GSN}\\  \textbf{Mean(N)}\end{tabular}} & \multicolumn{1}{c}{\begin{tabular}[c]{@{}c@{}}\textbf{M-GSN}\\ \textbf{Sim}\end{tabular}} & \multicolumn{1}{c}{\begin{tabular}[c]{@{}c@{}}\textbf{M-GSN}\\  \textbf{Sim(N)}\end{tabular}} \\ 
\midrule
\multirow{8}{*}{IMDB}        & \multirow{4}{*}{\begin{tabular}[c]{@{}l@{}}Macro F1\end{tabular}} & 20\%                         & 52.73                   & 53.64                   & 56.19                   & 59.35                                                                         & 58.41                                                                         & \textbf{61.04}                                                               & 60.67                                                                           & \textbf{61.01}                                                              & \textbf{60.74}                                                                 \\
                             &                                                                         & 40\%                         & 53.67                   & 55.50                   & 56.15                   & 60.27                                                                         & 59.44                                                                         & 61.28                                                                        & \textbf{61.36}                                                                  & \textbf{61.34}                                                              & \textbf{61.36}                                                                 \\
                             &                                                                         & 60\%                         & 54.24                   & 56.46                   & 57.29                   & 60.66                                                                         & 59.91                                                                         & 61.47                                                                        & \textbf{61.50}                                                                  & \textbf{61.59}                                                              & \textbf{61.48}                                                                 \\
                             &                                                                         & 80\%                         & 54.77                   & 57.43                   & 58.51                   & 61.44                                                                         & 60.20                                                                         & \textbf{61.92}                                                               & \textbf{61.82}                                                                  & \textbf{62.04}                                                              & 61.81                                                                          \\ \cline{2-12} 
                             & \multirow{4}{*}{\begin{tabular}[c]{@{}l@{}}Micro  F1\end{tabular}} & 20\%                         & 52.80                   & 53.64                   & 56.32                   & 59.60                                                                         & 58.42                                                                         & \textbf{61.08}                                                               & 60.81                                                                           & \textbf{61.03}                                                              & \textbf{60.87}                                                                 \\
                             &                                                                         & 40\%                         & 53.76                   & 55.56                   & 57.32                   & 60.50                                                                         & 59.51                                                                         & 61.40                                                                        & \textbf{61.52}                                                                  & \textbf{61.44}                                                              & \textbf{61.53}                                                                 \\
                             &                                                                         & 60\%                         & 54.23                   & 56.47                   & 58.42                   & 60.88                                                                         & 59.96                                                                         & 61.55                                                                        & \textbf{61.65}                                                                  & \textbf{61.66}                                                              & \textbf{61.61}                                                                 \\
                             &                                                                         & 80\%                         & 54.63                   & 57.40                   & 59.24                   & 61.53                                                                         & 60.28                                                                         & \textbf{62.01}                                                               & \textbf{61.98}                                                                  & \textbf{62.14}                                                              & 61.97                                                                          \\
                             \midrule
\multirow{8}{*}{DBLP}        & \multirow{4}{*}{\begin{tabular}[c]{@{}l@{}}Macro F1\end{tabular}} & 20\%                         & 88.00                   & 91.05                   & 91.69                   & 93.13                                                                         & 92.61                                                                         & \textbf{93.33}                                                               & \textbf{93.78}                                                                  & 92.95                                                                       & \textbf{93.84}                                                                 \\
                             &                                                                         & 40\%                         & 89.00                   & 91.24                   & 91.96                   & 93.23                                                                         & 93.04                                                                         & \textbf{93.49}                                                               & \textbf{93.84}                                                                  & 93.11                                                                       & \textbf{93.92}                                                                 \\
                             &                                                                         & 60\%                         & 89.43                   & 91.42                   & 92.14                   & 93.57                                                                         & 93.23                                                                         & \textbf{93.60}                                                               & \textbf{93.93}                                                                  & 93.32                                                                       & \textbf{93.96}                                                                 \\
                             &                                                                         & 80\%                         & 89.98                   & 91.73                   & 92.50                   & \textbf{94.10}                                                                & 93.38                                                                         & 93.59                                                                        & \textbf{93.90}                                                                  & 93.40                                                                       & \textbf{93.99}                                                                 \\ \cline{2-12}
                             & \multirow{4}{*}{\begin{tabular}[c]{@{}l@{}}Micro F1\end{tabular}} & 20\%                         & 88.51                   & 91.61                   & 92.33                   & 93.61                                                                         & 93.16                                                                         & \textbf{93.82}                                                               & \textbf{94.22}                                                                  & 93.48                                                                       & \textbf{94.28}                                                                 \\
                             &                                                                         & 40\%                         & 89.22                   & 91.77                   & 92.57                   & 93.68                                                                         & 93.53                                                                         & \textbf{93.95}                                                               & \textbf{94.25}                                                                  & 93.60                                                                       & \textbf{94.32}                                                                 \\
                             &                                                                         & 60\%                         & 89.57                   & 91.97                   & 92.72                   & 93.99                                                                         & 93.73                                                                         & \textbf{94.08}                                                               & \textbf{94.35}                                                                  & 93.82                                                                       & \textbf{94.38}                                                                 \\
                             &                                                                         & 80\%                         & 90.33                   & 92.24                   & 93.23                   & \textbf{94.47}                                                                & 93.85                                                                         & 94.04                                                                        & \textbf{94.31}                                                                  & 93.86                                                                       & \textbf{94.38}        
                             \\ \bottomrule                                                        
\end{tabular}}
\end{table*}

For GNNs, including GCN, GAT, HAN, MAGNN and our proposed M-GSN, we set the dropout rate to 0.5; we use the same splits of train/val/test sets; we employ the Adam optimizer with the learning rate set to 0.005 and the weight decay set to 0.001; we train the GNNs for 100 epochs and apply early stopping with a patience of 30. For node classification, the GNNs are trained in a semi-supervised fashion with a small fraction of nodes labeled as guidance. For a fair comparison, we set the embedding dimension of all the models mentioned above to 64. In the table \ref{tb:mgsn}, MAGNN paper is the result reported in \citep{fu2020magnn}, MAGNN repro is the result of our reproducing the MAGNN algorithm, M-GSN Mean represents the use of MEAN aggregation, M-GSN Mean(N) represents the use of MEAN aggregation while adding the layer L2 normalization, M-GSN Sim represents using SIM aggregation, and M-GSN Sim(N) represents the use of SIM aggregation while adding the layer L2 normalization. In each row of results, we highlight the Top 3 of the F1 metric. From the results, we can find that in these GNN algorithms, our M-GSN basically occupies the Top 3 of the F1 metric, which shows the effectiveness of the M-GSN algorithm. In the node classification task, our algorithm can learn a better node representation of the node.

\section{Conclusion}

In this paper, we define a general heterogeneous message passing paradigm, under which we can design different heterogeneous graph neural network models. In theory, any homogenous graph neural network can be migrated to this framework to obtain the corresponding heterogeneous version (homogeneous graph aggregation is equivalent to Intra-relation aggregation in this paradigm). The R-GSN we designed under this paradigm can achieve the state-of-the-art performance on the $\texttt{ogbn-mag}$ dataset. Of course, R-GSN can still be further improved. For example, our subgraph sampling algorithm is simple neighborhood sampling, we only experimented with 2 layers of R-GSN, which is limited by the 11GB memory limit of our NVIDIA-1080Ti GPU. The aggregation method of GSN is still the first-order neighborhood aggregation, therefore, our R-GSN can only receive the information of 2-hop neighbor nodes, and cannot receive the information of semantic high-order neighbors. Deepening the number of layers while avoiding overfitting is still an open question.

\bibliography{example_paper}
\bibliographystyle{icml2021}


\end{document}